# A New Clustering Approach based on Page's Path Similarity for Navigation Patterns Mining

**Heidar Mamosian**
Department of Computer Engineering, Science and Research Branch, Islamic Azad University (IAU), Khouzestan, Iran .

**Amir Masoud Rahmani**
Department of Computer Engineering, Science and Research Branch, Islamic Azad University (IAU),Tehran, Iran
.

**Mashalla Abbasi Dezfouli**
Department of Computer Engineering, Science and Research Branch, Islamic Azad University (IAU), Khouzestan, Iran .

*Abstract*—In recent years, predicting the user's next request in web navigation has received much attention. An information source to be used for dealing with such problem is the left information by the previous web users stored at the web access log on the web servers. Purposed systems for this problem work based on this idea that if a large number of web users request specific pages of a website on a given session, it can be concluded that these pages are satisfying similar information needs, and therefore they are conceptually related. In this study, a new clustering approach is introduced that employs logical path storing of a website pages as another parameter which is regarded as a similarity parameter and conceptual relation between web pages. The results of simulation have shown that the proposed approach is more than others precise in determining the clusters.

## I. INTRODUCTION

As the e-business is booming along with web services and information systems on the web, it goes without saying that if a website cannot respond a user's information needs in a short time, the user simply refers to another website. Since websites live on their users and their number, predicting information needs of a website's users is essential, and therefore it has gained much attention by many organization and scholars. One important source which is useful in analyzing and modeling the users' behavior is the second-hand information left by the previous web users. When a web user visits a website, for one request ordered by the user one or more than one record(s) of the server is stored in the web access log. The analysis of such data can be used to understand the users' preferences and behavior in a process commonly referred to as Web Usage Mining (WUM) [1, 2].

Most WUM projects try to arrive at the best architecture and improve clustering approach so that they can provide a better model for web navigation behavior. With an eye to the hypotheses of Visit-Coherence, they attempt to achieve more precise navigation patterns through navigation of previous web users and modeling them. As such, the prediction system on large websites can be initiated only when firstly web access logs are numerous. In other words, for a long time a website should run without such system to collect such web access log, and thereby many chances of the website are missed. Secondly, those involved in designing the website are not consulted.

Website developers usually take pages with related content and store them in different directories hierarchically. In this study, such method is combined with collected information from previous web users' navigation to introduce a new approach for pages clustering. The simulation results indicated that this method enjoys high accuracy on prediction. The rest of paper is structured as follows: section II outlines general principles. Section III described related work, and section 4 elaborates on a new clustering approach based on pages storage path. Section 5 reports on the results and section 6 is devoted to conclusion and future studies.

## II. PRINCIPLES

### A. Web Usage Mining process

Web usage mining refers to a process where users' access patterns on a website are studied. In general it is consists of 8 steps [3, 4]:

- Data collection. This is done mostly by the web servers; however there exist methods, where client side data are collected as well.

- Data cleaning. As in all knowledge discovery processes, in web usage mining can also be happen that such data is recorded in the log file that is not useful for the further process, or even misleading or faulty. These records have to be corrected or removed.

- User identification. In this step the unique users are distinguished, and as a result, the different users are identified. This can be done in various ways like using IP addresses, cookies, direct authentication and so on.

- Session identification. A session is understood as a sequence of activities performed by a user when he is navigating through a given site. To identify the sessions from the raw data is a complex step, because the server logs do not always contain all the information needed. There are Web server logs that do not contain enough information to reconstruct the user sessions; in this case (for example time-oriented or structure-oriented) heuristics can be used.

- Feature selection. In this step only those fields are selected, that are relevant for further processing.

- Data transformation, where the data is transformed in such a way that the data mining task can use it. For example strings are converted into integers, or date fields are truncated etc.

- Executing the data mining task. This can be for example frequent itemset mining, sequence mining, graph mining, clustering and so on.

- Result understanding and visualization. Last step involves representing knowledge achieved from web usage mining in an appropriate form.

As it was shown, the main steps of a web usage mining process are very similar to the steps of a traditional knowledge discovery process.

*B. Web Access Log*

The template is used to format your paper and style the text. All margins, column widths, line spaces, and text fonts are prescribed; please do not alter them. You may note peculiarities. For example, the head margin in this template measures proportionately more than is customary. This measurement and others are deliberate, using specifications that anticipate your paper as one part of the entire proceedings, and not as an independent document. Please do not revise any of the current designations.

Each access to a Web page is recorded in the web access log of web server that hosts it. Each entry of web access log file consists of fields that follow a predefined format. The fields of the common log format are [5]:

*remothost rfc931 authuser date "request" status bytes*

In the following a short description is provided for each field:

- **remotehost**. Name of the computer by which a user is connected to a web site. In case the name of computer is not present on DNS server, instead the computer's IP address is recorded.

- **rfc931**. The remote log name of the user.

- **authuser**. The username as witch the user has authenticated himself, available when using password protected WWW pages.

- **date**. The date and time of the request.

- **request**. The request line exactly as it came from the client (the file, the name and the method used to retrieve it).

- **status**. The HTTP status code returned to the client, indicating whether or not the file was successfully retrieved and if not, what error message was returned.

- **Byte**. The content-length of the documents transferred.

W3C presented an improved format for Web access log files, called extended log format, partially motivated by the need to support collection of data for demographic analysis and for log summaries. This format permits customized log files to be recorded in a format readable by generic analysis tools. The main extension to the common log format is that a number of fields are added to it. The most important are: *referrer*, which is the URL the client was visiting before requesting that URL, *user_agent*, which is the software the client claims to be using and *cookie*, in the case the site visited uses cookies.

III. RELATED WORK

In recent years, several Web usage mining projects have been proposed for mining user navigation behavior [6-11]. PageGather (Perkowitz, et al. 2000) is a web usage mining system that builds index pages containing links to pages similar among themselves. Page Gather finds clusters of pages. Starting from the user activity sessions, the co-occurrence matrix M is built. The element $M_{ij}$ of M is defined as the conditional probability that page *i* is visited during a session if page *j* is visited in the same session. From the matrix M, The undirected graph G whose nodes are the pages and whose edges are the non-null elements of M is built. To limit the number of edges in such a graph a threshold filter specified by the parameter MinFreq is applied. Elements of $M_{ij}$ whose value is less than MinFreq are too little correlated and thus discarded. The directed acyclic graph G is then partitioned finding the graph's cliques. Finally, cliques are merged to originate the clusters.

One important concept introduced in [6] is the hypotheses that users behave coherently during their navigation, i.e. pages within the same session are in general conceptually related. This assumption is called *visit coherence*.

Baraglia and Palmerini proposed a WUM system called SUGGEST, that provide useful information to make easier the web user navigation and to optimize the web server performance [8-9]. SUGGEST adopts a two levels architecture composed by an offline creation of historical knowledge and an online engine that understands user's behavior. In this system, PageGather clustering method is employed, but the co-occurrence matrix elements are calculated according to (1):

$$Mij = \frac{Nij}{\max(Ni, Nj)} \quad (1)$$

Where $N_{ij}$ is the number of sessions containing both pages i and j, $N_i$ and $N_j$ are the number of sessions containing only page i or j, respectively. Dividing by the maximum between single occurrences of the two pages has the effect of reducing the relative importance of index pages.

In SUGGEST a method is presented to quantify intrinsic coherence index of sessions based on visit coherence hypothesis. It measures the percentage of pages inside a user session which belong to the cluster representing the session considered. To calculate this index, the datasets obtained from the pre-processing phase is divided into two halves, apply the Clustering on one half and measure visit-coherence criterion on the basis of the second half. It is calculated according to achieved clusters. Measure of γ for each session is calculated according to 2:

$$\gamma_i = \frac{|\{p \in s_i \mid p \in C_i\}|}{N_i} \quad (2)$$

Where *p* is a page, *Si* is *i*-th session, *Ci* is the cluster representing *i*, and *Ni* is the number of pages in *i*-th session. The average value for γ over all $N_S$ sessions contained inside the dataset partition treated is given by:

$$\Gamma = \frac{\sum_{i=1}^{N_s} \gamma_i}{N_s} \quad (3)$$

Jalali et al. [10,11] proposed a recommender system for navigation pattern mining through Web usage mining to predict user future movements. The approach is based on the graph partitioning clustering algorithm to model user navigation patterns for the navigation patterns mining phase.

All these works attempted to find an architecture and algorithm for enhancing clustering quality, but the quality of achieved clusters is still far from satisfying. In this work, a clustering approach is introduced that is based on path similarity of web pages to enhance clustering accuracy.

## IV. SYSTEM DESIGN

The proposed system aims at presenting a useful information extraction system from web access log files of web servers and using them to achieve clusters from related pages in order to help web users in their web navigation. Our system has two modules. The pre-processing module and the module of navigation pattern mining. Figure 2 illustrate the model of the system.

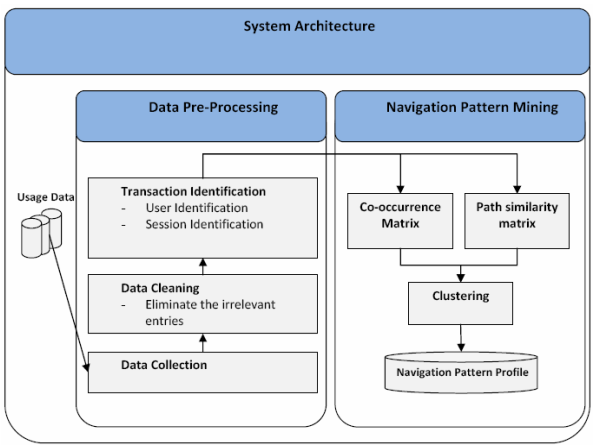

Figure 1.    Model of the system.

### A. Data Pre-processing

There are several tasks in data pre-processing module. We begin by removing all the uninteresting entries from the web access log file which captured by web server, supposed to be in Common Log Format. Namely, we remove all the non-html requests, like images, icons, sound files and generally multimedia files, and also entries corresponding to CGI scripts. Also the dumb scans of the entire site coming from robot-like agents are removed. We used the technique described in [12] to model robots behavior.

Then we create user sessions by identifying users with their IP address and sessions by means of a predefined timeout between two subsequent requests from the same user. According to Catledge et al. in [13] we fixed a timeout value equal to 30 minutes.

### B. Navigation pattern mining

After the data pre-processing step, we perform navigation pattern mining on the derived user access session. As an important operation of navigation pattern mining, clustering aims to group sessions into clusters based on their common properties. Here, to find clusters of correlated pages, both website developers and website users are consulted. To do so, two matrixes M and P are created. Matrix M is a co-occurrence matrix which represents website users' opinions, and matrix P is the matrix of pages' path similarity.

*1) Co-occurrence Matrix:* The algorithm introduced at SUGGEST system [8, 9] is employed to create co-occurrence matrix. Using this algorithm, M co-occurrence matrix is created which represents corresponding graph with previous users of a website. The elements of this matrix are calculated based on (1) appeared in section III.

*2) Path similarity matrix:* Website developers usually store pages which are related both in structure and content is same subdirectory, or create links between two related pages. Due to our lack of knowledge about links between pages on web access logs, to realize the developer's opinion on conceptual relation between pages, the website's pages storage path is employed. For example, two pages *Pi* and *Pj* which are located in the following paths.

*Directory1/Subdir1/subdir2/p1.html*
*Directory1/Subdir1/subdir2/p2.html*

Are more related than two pages which are on the following paths

*Directory1/Subdir1/subdir2/p1.html*
*Directory2/Subdir3/subdir4/p2.html*

Hence, a new matrix called pages' path similarity matrix can be achieved. To calculate path similarity matrix, first the function *similarity(Pi , Pj)* is defined. This function returns the number of common sub-directories of two pages, i.e. *Pi* and *Pj*. To calculate path similarity matrix elements, the following equation is used:

$$P_{ij} = \frac{2 \times similarity(path(p1), path(p2))}{number\ of\ directory(path(p1)) + number\ of\ directory(path(p1))} \quad (4)$$

Where number of directory(path(Pi)) is the number of sub-directories of storage path in Pi. When two paths of two pages are close to each other, the value of each element of this matrix get closer to 1, and if there is no similarity in storage path, it becomes 0.

**Example**: For two pages, i.e. p1.html and p2.html which are stored on the following paths:

Pi:/history/skylab/pi.html

Pj: /history/mercury/ma8/pj.html

Then

$$P_{ij} = \frac{2 \times 1}{2+3} = 0.4$$

*3) Clustering Algorithm:* Combining these two matrixes, the new matrix C is created which shows relation between different pages of site based on a mix of users and developers opinions. To combine these two matrixes whose elements of each varies between zero and 1, Equation (5) is used to keep the values of combined matrixes still between zero and 1.

$$C_{ij} = \alpha \times M_{ij} + (1-\alpha) \times P_{ij} \qquad (5)$$

Where M is co-occurrence matrix and P is the path similarity matrix. To arrive at clusters of related pages, the graph corresponding to the achieved matrix is divided into strong partitions. To do this, DFS algorithm is employed as follows. When the value of *Cij* is higher than the *MinFreq*, two corresponding nodes are considered connected, and in other case they are taken disconnected. We start from one node and find all nodes connected to it through execution of DFS algorithm and put them on one cluster. Each node which is visited is labeled with a visited label. If all nodes bear visited labels, the algorithm ends, otherwise the node not visited is selected and DFS algorithm id performed on it, and so on.

## V. EXPERIMENTAL EVALUATION

For an experimental evaluation of the presented system, a web access log from NASA produced by the web servers of Kennedy Center Space. Data are stored according to the Common Log Format. The characteristics of the dataset we used are given in Table 1.

TABLE I. DATASET USED IN THE EXPERIMENTS.

| Dataset | Size(MB) | Records(thousands) | Period(days) |
|---------|----------|--------------------|--------------|
| NASA    | 20       | 1494               | 28           |

All evaluation tests were run on Intel® Core™ Duo 2.4 GHz with 2GB RAM, operating system Windows XP. Our implementation have run on .Net framework 3.5 and VB.Net and MSSqlServer 2008 have used for coding the proposed system.

TABLE II. REMOVED EXTRA ENTRIES

| Page Extension | Count of Web Log Entries |
|----------------|--------------------------|
| .gif           | 899,883                  |
| .xbm           | 43,954                   |
| .pl            | 27,597                   |
| .bmp, .wav, …, web bots entries | 165,459     |
| **Total**      | **1,136,893**            |

After removing extra entries, different web users are identified. This step is conducted based on remotehost field. After identified distinct web users, users' sessions are reconstructed. As sessions with one page length are free from any useful information, they are removed too. In Table 3, characteristics of web access log file is represented after performing pre-processing phase.

TABLE III. CHARACTERISTICS OF WEB ACCESS LOG FILE AFTER PERFORMING PRE-PROCESSING PHASE

| Dataset | Size(MB) | Number of Records | Number of Distinct Users | Number of Sessions |
|---------|----------|-------------------|--------------------------|--------------------|
| NASA    | 20       | 357,621           | 42,215                   | 69,066             |

As shown in Figure 3, the percentage of sessions formed by a predefined number of pages quickly decreases when the minimum number of pages in a session increases.

First all the uninteresting entries from the web access log file (entries corresponding to multimedia logs, CGI scripts and corresponding inputs through navigations of web bots) are removed.

For example, samples of these extra inputs are cited in Table 2 along with the number of their repetition in NASA's web access log.

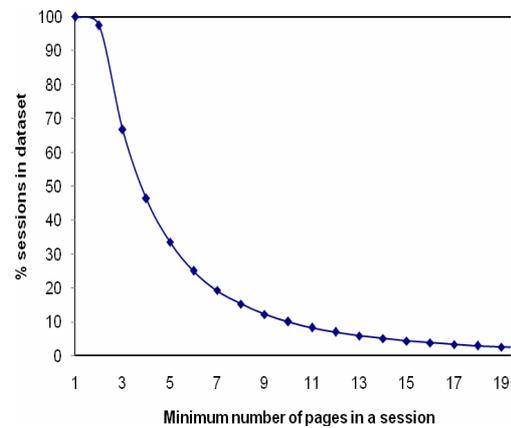

Figure 2. Minimum number of pages in session.

Once the users sessions are reconstructed based on clustering algorithm presented in section 4.2.3, clustering operation is calculated based on varying values of MinFreq and α, the percentage of pages clustered is calculated. The tests showed that the percentage of participated pages for value α = 0.8 is at the best status. In Figure 4, the percentage of clustered pages is represented as a function of the parameter *MinFreq* and for two values α = 1.0 and α =0.8.

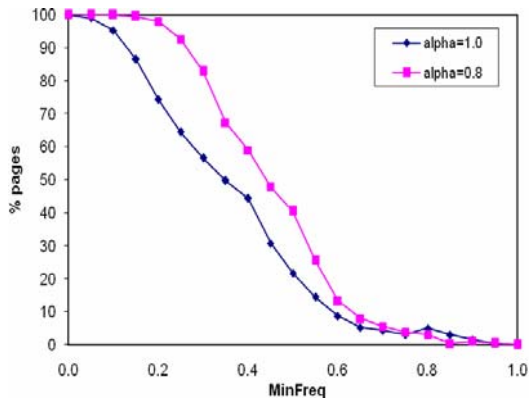

Figure 3. Minimum number of pages in session.

Figure 5 show the number of achieved clusters for two values α=1.0 and α=0.8 as a function of the MinFreq parameter.

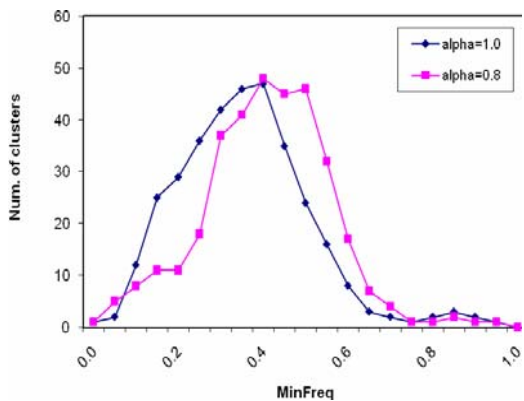

Figure 4. Number of Clusters Found.

To evaluate the quality of clusters found for varying values of α, the Visit-coherence index introduced in section 2 is employed.

In Figure 6, the value of ґ, is represented as a function of the MinFreq parameter for two values α. As shown in the figure 6, using our proposed clustering algorithm has enhanced clusters' quality.

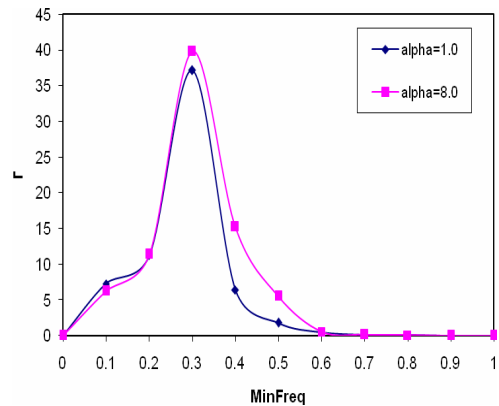

Figure 5. Coherence of visit

## VI. CONCLUSION

For an experimental evaluation of the presented system, a web access log from NASA produced by the web servers of Kennedy Center Space. Data are stored according to the Common Log Format. The characteristics of the dataset we used are given in